\documentclass[letterpaper, 10 pt, conference]{ieeeconf}  

\IEEEoverridecommandlockouts                              

\usepackage{cite}
\usepackage{graphics} 
\usepackage{epsfig} 
\usepackage{mathptmx} 
\usepackage{times} 
\usepackage{amsmath} 
\usepackage{amssymb}  
\usepackage{multirow}
\usepackage{color}
\usepackage{xcolor}
\usepackage{colortbl}
\usepackage{tabulary}
\usepackage[normalem]{ulem}
\usepackage{makecell}
\usepackage{algpseudocode}
\usepackage{algorithm}

\title{\LARGE \bf
6D Robotic Assembly Based on RGB-only Object Pose Estimation
}

\author{Bowen Fu$^{*}$, Sek Kun Leong$^{*}$, Xiaocong Lian and Xiangyang Ji 
\thanks{$^{*}$ Equal contribution}
\thanks{This work was supported by the National Key R\&D Program of China under Grant 2018AAA0102801, National Natural Science Foundation of China under Grant 61620106005.}
\thanks{All authors are with the Department of Automation and BNRist, Tsinghua University, Beijing, China. 
        {\tt\small \{fbw19, lxq20\}@mails.tsinghua.edu.cn, \{lian900625, xyji\}@tsinghua.edu.cn}
}}

\begin{document}

\maketitle
\thispagestyle{empty}
\pagestyle{empty}

\begin{abstract}

Vision-based robotic assembly is a crucial yet challenging task as the interaction with multiple objects requires high levels of precision. 
In this paper, we propose an integrated 6D robotic system to perceive, grasp, manipulate and assemble blocks with tight tolerances. 
Aiming to provide an off-the-shelf RGB-only solution, our system is built upon a monocular 6D object pose estimation network trained solely with synthetic images leveraging physically-based rendering.
Subsequently, pose-guided 6D transformation along with collision-free assembly is proposed to construct any designed structure with arbitrary initial poses. 
Our novel 3-axis calibration operation further enhances the precision and robustness by disentangling 6D pose estimation and robotic assembly. 
Both quantitative and qualitative results demonstrate the effectiveness of our proposed 6D robotic assembly system. 

\end{abstract}
\section{Introduction}

\begin{figure}
	\centering
	\vspace{2mm}
	\includegraphics[width = 0.95\linewidth]{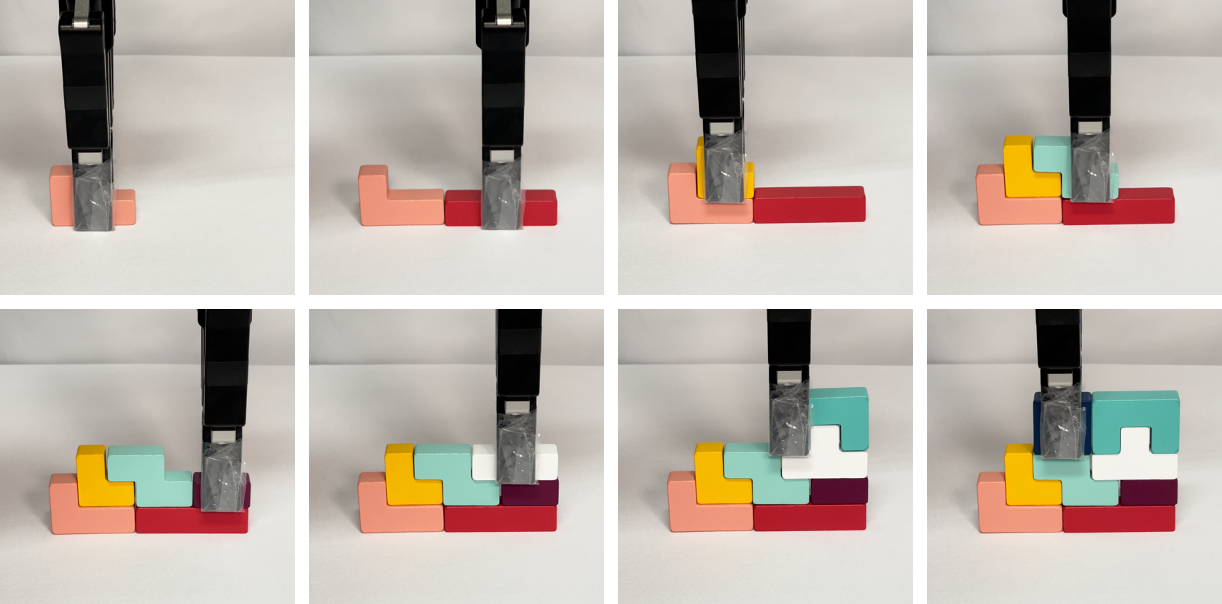}\vspace{-2mm}
	\caption{Example of an assembly process. Given a desired architecture, our system autonomously and precisely assembles the blocks. }
	\label{fig:assemble}
\end{figure}

Although building blocks is natural for humans, it is quite challenging for robots. 
A block needs to be perceived, grasped, manipulated, and then appropriately assembled with an extremely tight tolerance, thus requiring a highly robust and precise vision algorithm. 
In this work, we focus on establishing a flexible 6D robotic system to assemble blocks based on monocular 6D object pose estimation (Fig.~\ref{fig:assemble}). 

Robotic assembly tasks including peg-in-hole and block stacking have been studied for decades. 
Peg-in-hole tasks are typically implemented on a board \cite{kimble2020benchmarking,yokokohji2019assembly,von2020robots}, thus only 2D information is required. 
Some works exploit 2D object detection \cite{triyonoputro2019quickly,almaghout2021robotic} or 3D pose estimation \cite{litvak2019learning} to perceive the objects. 
By simplifying the assembly task from 3D space into a 2D plane with only 2 or 3 degrees of freedom, the information along the principal axis is lost.
Therefore, only unidirectional assembly can be implemented, rendering flexible multi-angle block assembly tasks impossible. 
Additionally, force-torque sensors are employed in some works \cite{furrer2017autonomous,fakhurldeen2019cara,almaghout2021robotic}, whereas the high cost oftentimes prohibits their practical applications. 

Some works leverage reinforcement learning to conduct block stacking tasks \cite{Zhu-RSS-18,lee2021beyond}, which attempt to teach robots to stack one block on top of another one. 
The learned policies are capable of dealing with multiple object combinations and demonstrate a large variety of stacking skills. 
However, only simple stacking operation can be implemented. 
There is still a long way to meet the requirement of robotic assembly. 

To conduct more complex grasping and assembly tasks, a robot needs to interact with objects in 3D space with full 6 degrees of freedom.
Thanks to the rapid development in convolutional neural networks (CNNs), recently, a few methods attempt to conduct robotic assembly through learning-based 6D object pose estimation. 
For instance, \cite{stevvsic2020learning} formulates the task of assembly as predicting the 6D poses of template geometries to enable manipulating objects in arbitrary contexts.
However, the object is manually placed into the gripper to avoid the grasp error accumulating to the assembly process, which lacks system integrity. 
Additionally, \cite{morgan_vision-driven_2021} conducts multiple open-world assembly tasks leveraging an RGBD-based 6D object pose tracker. 
However, the dependence on surrounding objects and depth information may restrict its application. 

In this work, an integrated robotic system is established to autonomously and precisely assemble blocks with tight tolerances. 
Only RGB input is required and the prior information of the object is strictly confined to an easily acquired 3D model, making it easy to extend to real-world applications. 
Our system is built upon a high-precision real-time monocular 6D object pose estimation methodology, with all the training and validation data generated by physically-based rendering.
The learned model can be directly applied to real-world robotic manipulation without further training or using domain randomization techniques. 
Aided by the proposed pose-guided 6D transformation strategy, our system is capable of assembling arbitrary given structures with arbitrary initial block poses. 
To further enhance the precision and robustness, 3-axis calibration is introduced to decouple 6D pose estimation and the robotic assembly process, which eliminates the effect of pose error on assembly. 

Our contributions can be summarized as follows: 

\begin{itemize}
\item We establish an integrated 6D robotic assembly system to assemble blocks with arbitrary initial poses to any given structure, only requiring RGB input. 
\item We propose a pose-guided 6D transformation strategy along with an RGB-based 6D pose regression network \cite{wang2021gdr} with pure synthetic training, providing high-precision real-time interaction between blocks and the robot system. 
\item We further promote the precision and robustness by 3-axis calibration. The effectiveness of our methodology is demonstrated by robotic assembly tasks with 1mm tolerance. 
\end{itemize}
\section{Related Work}

The presented work relates to two major strands of research: 6D object pose estimation and vision-based robotic assembly. 

\subsection{6D Object Pose Estimation}

6D object pose estimation has received a lot of attention in both robotics and computer vision communities in recent years. 
Some works conduct indirect approaches to predict the 6D pose. 
A popular approach is to establish 2D-3D correspondences, which are subsequently exploited for computing the 6D poses by P$n$P algorithm. 
For instance, BB8 \cite{rad2017bb8} and YOLO6D \cite{tekin2018real} compute the 2D projections of 3D bounding box corners. 
To enhance the robustness, SegDriven \cite{hu2019segmentation} and PVNet \cite{peng2019pvnet} employ segmentation paired with voting for each correspondence. 
Meanwhile, CDPN~\cite{li2019cdpn} and DPOD \cite{zakharov2019dpod} predict dense rather than sparse correspondences and achieve remarkable performance. 
Similarly, EPOS \cite{hodan2020epos} represents the object by compact surface fragments to handle symmetries. 

Another branch of works directly regresses the 6D pose. 
PoseCNN \cite{xiang2018posecnn} and DeepIM \cite{li2018deepim} leverage a point matching loss and CosyPose \cite{labbe2020cosypose} extends DeepIM \cite{li2018deepim} by introducing multi-view information and implementing global scene refinement. 
SingleStage \cite{hu2020single} and GDR-Net \cite{wang2021gdr} infer intermediate 2D-3D correspondences and directly regress the 6D pose via the network rather than the P$n$P algorithm, showing that the learned P$n$P is capable of producing more robust estimates than standard P$n$P. 
Additionally, SSD6D \cite{kehl2017ssd} discretizes the pose space and conducts classification rather than regression. 

Noteworthy, the majority of these methods exploit annotated real data for training. 
However, labeling real data requires tremendous consumption of time and labor. 
Hence, some works completely rely on synthetic data to avoid this defect \cite{sundermeyer2018implicit,manhardt2019explaining}. 
Though the performance of these methods falls behind those using real annotations, the domain gap between synthetic and real images can be narrowed by physically-based rendering \cite{hodavn2020bop, denninger2020blenderproc} or domain randomization strategies. 

\begin{figure*}
	\centering
	\vspace{2mm}
	\includegraphics[width = 0.99\linewidth]{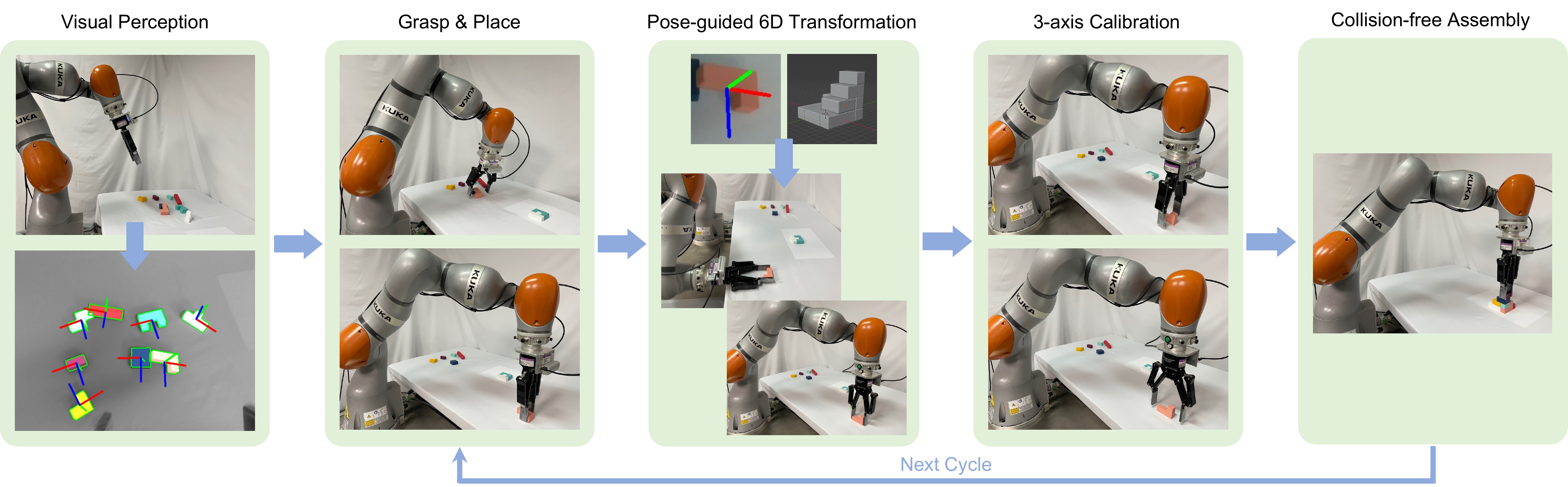}\vspace{-2mm}
	\caption{Flow chart of our system. After perceiving all the blocks by 2D object detection and 6D object pose estimation, we implement grasp \& place, pose-guided 6D transformation, 3-axis calibration and collision-free assembly for each block until the intact architecture is constructed. }
	\label{fig:flow_chart}
\end{figure*}

\subsection{Vision-based Robotic Assembly}

With the blossom of deep learning, vision-based robotic grasping and assembly progress rapidly. 
Compared with robotic grasping tasks~\cite{loing2018virtual,wen2020se}, robotic assembly tasks require higher precision because the grasped object interacts not only with the gripper but also with the target object. 
Peg-in-hole insertion tasks and block stacking tasks have drawn much attention among others. 

Peg-in-hole insertion tasks require robots to assemble gears, shafts, etc. 
Various challenges have been proposed in the last few years, \textit{e.g.}, the NIST Assembly Task Boards~\cite{kimble2020benchmarking} and the World Robot Summit (WRS) Assembly Challenge ~\cite{yokokohji2019assembly,von2020robots}.
The targets are typically placed on a plane, thus many peg-in-hole insertion methods leverage 2D object detection \cite{triyonoputro2019quickly} or 3D pose estimation (\textit{i.e.}, the 2D object center $(x,~y)$ and the orientation angle $\theta$) \cite{litvak2019learning}. 
\cite{almaghout2021robotic} utilizes YOLOv5 to detect target objects and accomplish assembly using visual and force servoing. 
Their interaction within the 3D space is limited and the flexible multi-angle assembly is beyond their reach. 

Meanwhile, several works deal with vision-based robot stacking tasks by reinforcement learning \cite{Zhu-RSS-18, lee2021beyond}, trying to teach robots to stack one block on top of the other block. 
Their methods leverage few prior and rely on physical simulation to perform large-scale robot training, which proves the possibility of learning a vision-based policy to stack multiple object combinations. 
However, only simple tasks, \textit{i.e.}, stack one block on top of the other, can be conducted.
There is still a long way towards high-precision block stacking tasks leveraging reinforcement learning. 

Recently, some works combine pose estimation and robotic assembly or stacking. 
\cite{stevvsic2020learning} introduces a new robotic assembly task, which requires reasoning about local geometry that is surrounded by arbitrary context and formulates the task as the 6D pose estimation of template geometries. 
\cite{morgan_vision-driven_2021} conducts plug insertion, box packing and cup stacking tasks leveraging RGBD-based 6D object pose tracking along with within-hand visual feedback control. 
\section{Method}

In this paper, we aim to exploit a 6D robotic assembly system to build blocks with tight tolerances by RGB-only input (Fig. \ref{fig:flow_chart}). 
To implement the task autonomously and precisely, we first perceive all the blocks leveraging 2D object detection and 6D object pose estimation. 
After conducting multiple grasp and place operations to interact with each block, pose-guided 6D transformation and collision-free assembly are proposed to deal with arbitrary initial poses and designed structures. 
Moreover, the key challenge of the assembly task is the gap between the precision and robustness required by the robot system and those provided by the computer vision algorithm. 
Therefore, 3-axis calibration is proposed to further enhance the precision and robustness by disentangling 6D pose estimation and the 6D assembly process. 

\subsection{Data Generation}

For training the 2D object detection network and the 6D object pose estimation network, we generate photorealistic images leveraging physically-based rendering. 
Only 3D object models are required in this process. 
Since the blocks are standard and regular, we manually model them by SolidWorks and re-sample the surface utilizing Poisson-disk sampling to obtain dense point clouds. 

Some 6D pose estimation methods \cite{li2018deepim,li2019cdpn,peng2019pvnet} render 3D object models in an arbitrary pose and randomly choose a background image from PASCAL VOC \cite{everingham2010pascal} or SUN397 \cite{xiao2010sun} dataset to generate synthetic data. 
However, the generated images are unrealistic due to the fragmented foreground and background, which triggers the domain gap between synthetic and real data. 

In contrast to them, we exploit BlenderProc \cite{denninger2020blenderproc}, a physically-based rendering pipeline, to generate synthetic training and validation data. 
Specifically, each object is assigned with a uniformly sampled initial pose and then dropped to the ground of a room with a random photorealistic material from the CC0 Textures library.
The synthetic images are rendered leveraging Blender with properly set light and camera positions. 
Since the Pybullet physics engine records the real-time object poses, we simultaneously acquire both 2D detection and 6D pose annotations of the images. 

As shown in Fig.~\ref{fig:PBR_real_pose}~(a), abundant high-quality images with precise annotations can be generated for any given model, which is highly efficient and labor-saving. 
Our methodology greatly narrows the domain gap between synthetic and real images. 
Subsequently, the 2D detection network and 6D pose estimation network, which are solely trained on synthetic data, can be directly applied to real scenes without further training or using domain randomization techniques. 

\subsection{6D Pose Estimation}
\begin{figure*}
	\centering
	\vspace{2mm}
	\includegraphics[width = 0.75\linewidth]{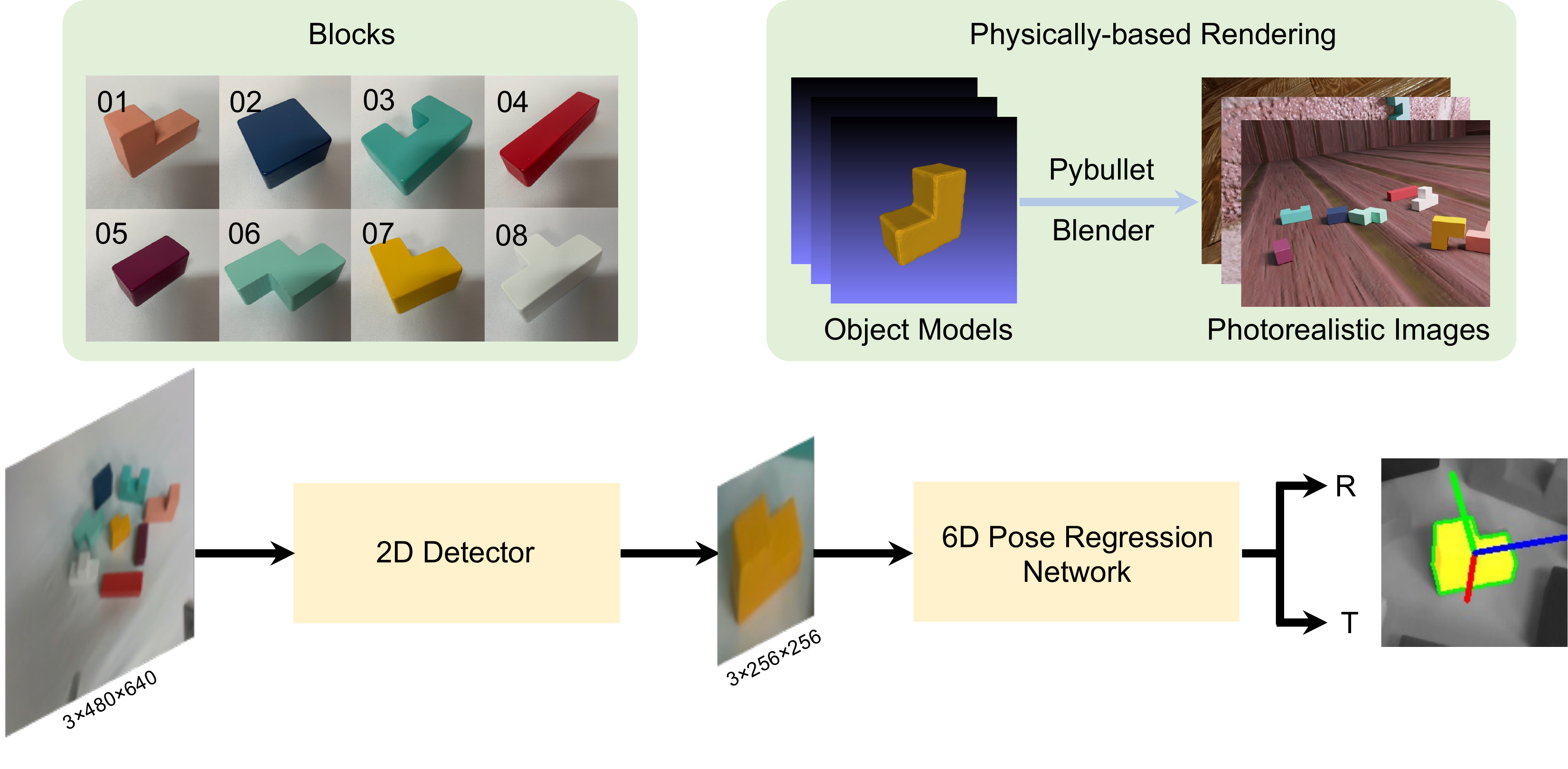}\vspace{-4mm}
	\caption{Overview of our 6D pose estimation methodology. We train the 2D detection network and the 6D pose estimation network with pure synthetic images leveraging physically-based rendering. During inference, we feed a real image captured by the camera to the networks and obtain the 6D pose. 8 blocks are employed in our experiments. }
	\label{fig:network}
\end{figure*}

Given an RGB image and a set of objects together with their corresponding 3D models, we aim to estimate the 6D pose $P = [R|t]$ w.r.t. the camera for each object, with $R$ representing the 3D rotation and $t$ denoting the 3D translation. 

Fig. \ref{fig:network} presents a schematic overview of the proposed methodology. 
We follow GDR-Net~\cite{wang2021gdr} to directly regress the 6D pose. 
Leveraging physically-based rendering, we train the 2D detection network and the 6D pose regression network solely on synthetic data. 
Specifically, we first detect all objects of interest using the trained detection network. 
For each corresponding Region of Interest (RoI), we feed it to the pose network exploiting Dynamic Zoom-In (DZI) \cite{li2019cdpn} strategy, which decouples the 2D detection process and 6D pose estimation process. 
Then several intermediate geometric feature maps indicating the 2D-3D correspondences are predicted, and finally, a CNN module that simulates the P$n$P algorithm is exploited to directly regress the 6D object pose. 

Following \cite{wang2021gdr}, we employ a disentangled 6D pose loss, individually supervising the rotation $R$, the scale-invariant 2D object center \cite{li2019cdpn} and the distance. 
We parameterize rotation $R$ as the first two columns of the rotation matrix, which has been demonstrated effective \cite{zhou2019continuity,wang2021gdr}. 

To sum up, with a glance of the camera mounted to the gripper, we acquire the current poses of all blocks w.r.t. the camera by our 6D pose estimation methodology. 
Aided by the high quality of training data and the generalization ability of the network, our system requires no prior of working scenario, which vastly extends its application range. 

\subsection{Grasp Strategy and Collision-free Assembly}

Based on the 6D poses we have acquired, the robot system is capable of interacting with the blocks. 
For implementing subsequent grasp operations, we first consider two coordinate systems: the object coordinate system $\mathnormal{O}_{obj}$, defined in the 3D object model, and the robot base coordinate system $\mathnormal{O}_{base}$. 
The 6D pose represents the transformation $T^{cam}_{obj}$ from  $\mathnormal{O}_{obj}$ to the camera coordinate system $\mathnormal{O}_{cam}$. 
With the offline calibration between the camera and the robot flange and the robot kinematics parameters, the transformation from $\mathnormal{O}_{obj}$ to $\mathnormal{O}_{base}$ is obtained by
\begin{equation}
T^{base}_{obj}=T^{base}_{flange}~T^{flange}_{cam}~T^{cam}_{obj},  
\end{equation}
where $T$ denotes the homogeneous transformation matrix. 
For simplification, we consider the axis of $\mathnormal{O}_{obj}$ with the minimum angle to the vertical axis of $\mathnormal{O}_{base}$ as the reference axis
\begin{equation}
I_{ref}=\arg\max_v \langle e_{z} ,~R^{base}_{obj} ~ v\rangle,  
\end{equation}
where $v$ denotes the unit vector of the axes of $\mathnormal{O}_{obj}$ and $e_z$ denotes the unit vector of the vertical axis of $\mathnormal{O}_{base}$. 

We conduct a grasp towards the block center along the reference axis to avoid the unreachable point of the robot arm and collisions with the other blocks. 
However, in the assembly process, a proper grasp position needs to be found for the gripper to avoid collisions with the other blocks. 
Considering the blocks are regular, we conduct grasping along the axes in $\mathnormal{O}_{obj}$, including $\pm x$, $\pm y$ and $\pm z$. 
For each direction, we pre-define two orthogonal grasps. 
For instance, the grasp along $+x$ includes two orthogonal grasps in $x$-$y$ plane and $x$-$z$ plane. 
Considering directly grasping the object center may be infeasible due to the collisions with the designed structure, we also pre-define two extra grasp positions for each direction, with offsets relative to the center. 
To sum up, 36 grasp candidates are pre-defined for each block in $\mathnormal{O}_{obj}$. 

Knowing the block poses and the grasp candidates, we detect collisions between the gripper and the designed structure leveraging the Flexible Collision Library (FCL)~\cite{pan2012fcl} to exclude infeasible grasp candidates. 
Noteworthy, the 2-finger gripper can be formulated as two cuboids, which is enough for filtering out infeasible grasp candidates. 

\subsection{Pose-guided 6D Transformation}
We design a target structure manually in Blender, where the relative 6D poses between each block can be obtained. 
Simultaneously, the assembly sequence is also allocated. 
As long as we assign the absolute 6D pose of the first block in $\mathnormal{O}_{base}$, all the poses of blocks can be inferred, which makes our method easy to transfer to other desired structures. 

Therefore, the arbitrary initial and target block poses have been obtained respectively, which guides the robot to implement 6D transformation for each block. 
Similar to the grasping process, the rotation along the vertical axis of $\mathnormal{O}_{base}$ is easy to deal with while the rotation along the other two axes may be out of range due to the limited workspace of the robot arm.
We would like to avoid wide angle rotation along the horizontal axes of $\mathnormal{O}_{base}$.

If the target pose shares the same reference axis with the current pose, the assembly can be conducted simply by virtue of the rotation along the vertical axis of $\mathnormal{O}_{base}$. 
Otherwise, we move the block to a rotation workspace and rotate it along one horizontal axis of $\mathnormal{O}_{base}$ to make the target reference axis upturned. 
Noteworthy, up to two rotations are sufficient for any arbitrary conditions and the number of rotations is related to the angle between the reference axes of the current pose and the target pose. 
Note that to account for symmetries, we compute all correct poses by multiplying the estimated pose and rotation matrix of symmetry and select the most convenient transformation strategy. 

\begin{figure*}[ht]
	\centering
    \vspace{2mm}
	\includegraphics[width = 0.99\linewidth]{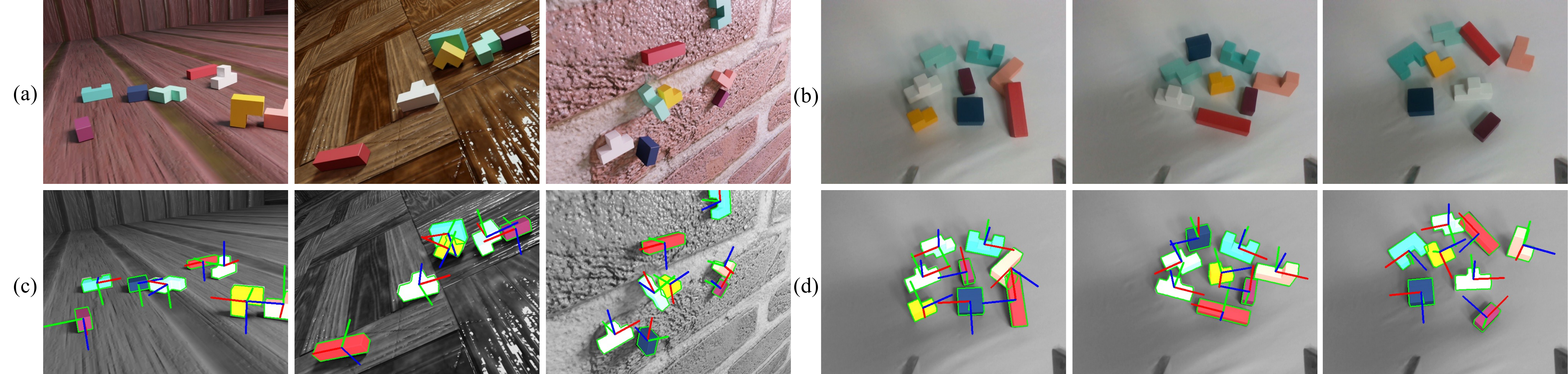}
	\vspace{-2mm}
    \caption{The qualitative 6D pose estimation results of photorealistic and real-time images. (a) The photorealistic images we generate leveraging physically-based rendering. (b) The real-time images we capture. (c,~d) The corresponding qualitative 6D pose estimation results by rendering the 3D models and overlaying the contours and coordinate axes on the image.}
	\label{fig:PBR_real_pose}
\end{figure*}

\subsection{3-axis Calibration}

Although we achieve relatively accurate 6D pose estimation, the assembly precision is still not satisfied. 
To be specific, in 6D pose estimation tasks, the average error of model points is usually considered acceptable up to 10\% of the diameter.
The grasp and place operation shares a similar tolerance and can still be implemented stably and robustly. 
However, in assembly tasks, the pose estimation error can lead to collisions or gaps. 
Considering the block only interacts with the gripper in the grasp and place operation but needs to interact with both the gripper and other blocks in the assembly operation, the precision required by the assembly operation is much higher than that required by the grasp and place operation. 
To eliminate the impact of pose error on assembly, we propose 3-axis calibration to decouple 6D pose estimation and assembly process. 

Assume that there is a small rotation error $ \Delta R $ and translation error $ \Delta t $ between ground-truth pose $ (\bar{R},~\bar{t}) $ and estimated pose $ (R,~t) $. 
We first grasp and place the block on a plane. 
Since the block is regular, $ \Delta R_x $, $ \Delta R_y $ and $ \Delta t_z $ can be eliminated due to the limitation of the plane. 
Then we conduct two orthogonal grasps along the \textit{x} axis and the \textit{y} axis of $\mathnormal{O}_{obj}$ to the estimated pose, forcing the block to shift to the estimated pose. 
Therefore, $ \Delta R_z $, $ \Delta t_x $ and $ \Delta t_y $ are eliminated, which are tiny yet crucial to assembly process. 

Noteworthy, 3-axis calibration shares the same plane prior with the preceding process, thus does not leverage supernumerary prior. 
On account of 3-axis calibration, the blocks can still be anywhere with arbitrary initial poses, even be stacked together. 
The pose error during perception, grasp and place process cannot be accumulated to the assembly process, thus the success rate is improved dramatically. 

\section{EXPERIMENTS}

We use eight blocks for 6D pose estimation and robotic assembly experiments. 
In this section, we first evaluate our 6D object pose estimation algorithm. 
Then robot assembly experiments are implemented to demonstrate the effectiveness of our methodology. 

\subsection{6D Pose Inference}

\subsubsection{Implementation Details} 
We generate 38000 photorealistic synthetic images for training and 2000 for validation. 
All eight blocks are dropped randomly from 0.2 to 0.4 m in a -0.15 to 0.15 m square with a uniformly sampled rotation matrix in Blender. 
The camera is placed within an annulus with 0.55 to 0.85 m radius and 0.5 to 0.9 m elevation. 
The 2D detection and 6D pose estimation experiments are implemented using PyTorch. 
We leverage YOLOv4 \cite{bochkovskiy2020yolov4} for detection. 
The pose network is trained end-to-end using Ranger optimizer with a batch size of 64 and a base learning rate of 1e-4. 

\begin{table}[t]
    \centering
    \caption{Results for 6D pose estimation on validation set}
    \label{tab:6D_pose_results}
    \setlength{\tabcolsep}{3.5mm}
    \begin{tabular}{cccccc}
    \hline
         \multirow{2}{*}{Block} & \multicolumn{3}{c}{ADD(-S)} & \multirow{2}{*}{5$^\circ$ 5cm} & \multirow{2}{*}{2cm}\\
                                 & 0.02d & 0.05d & 0.1d       & & \\
    \hline
         01 & 57.39 & 82.91 & 88.10 & 80.95 & 89.26\\
         02 & 51.28 & 81.47 & 88.59 & 53.53 & 91.22\\
         03 & 61.23 & 83.40 & 89.69 & 82.35 & 90.43\\
         04 & 54.50 & 79.81 & 87.69 & 73.81 & 87.81\\
         05 & 43.74 & 78.26 & 88.71 & 82.32 & 92.71\\
         06 & 60.83 & 85.33 & 90.61 & 82.75 & 91.16\\
         07 & 44.79 & 79.63 & 88.82 & 80.91 & 91.32\\
         08 & 53.17 & 83.96 & 90.11 & 84.39 & 91.79\\
    \hline
         Mean & 53.37 & 81.84 & 89.04 & 77.63 & 90.71\\
    \hline
    \end{tabular}
\end{table}

\subsubsection{Evaluation Metrics} 
We employ ADD(-S) and \textit{n}$^\circ$ \textit{n} cm for evaluation. 
The ADD metric~\cite{hinterstoisser2012model} measures whether the average distance of the model vertices between the predicted pose and the ground-truth pose is less than a certain percentage of the object's diameter. 
For symmetric objects, the ADD-S metric is employed to measure the error as the average distance to the closest model vertices \cite{hinterstoisser2012model}\cite{hodavn2016evaluation}. 
The \textit{n}$^\circ$ \textit{n} cm metric measures whether the rotation error is under \textit{n}$^\circ$ and the translation error is less than \textit{n} cm. 
It is computed w.r.t. the smallest error for all possible ground-truth poses for symmetric objects. 

\subsubsection{Analysis} 
Table~\ref{tab:6D_pose_results} shows the performance of our 6D pose estimation methodology. 
As can be seen, we achieve 89.04 in ADD(-S)-0.1d and 77.63 in 5$^\circ$ 5cm on the validation set. 
With regard to more rigorous metrics ADD(-S)-0.05d and ADD(-S)-0.02d, we achieve 81.84 and 53.37, respectively. 
We also demonstrate qualitative results in Fig.~\ref{fig:PBR_real_pose} for synthetic and real-world data. 
Although we do not acquire the ground-truth 6D poses in real scenes, the visualization reflects the high precision of our methodology. 
Aided by our 6D pose estimation methodology, arbitrary initial block poses can be handled even when the blocks are stacked together with occlusion. 

\subsubsection{Inference time} 
With an AMD Ryzen 7 5800 CPU and an NVIDIA RTX 3090 GPU, given a $640 \times 480$ image, using YOLOv4 \cite{bochkovskiy2020yolov4} detector, our approach takes 45ms for eight objects, including 21ms for detection. 

\begin{figure}
	\centering
	\vspace{2mm}
	\includegraphics[width = 0.95\linewidth]{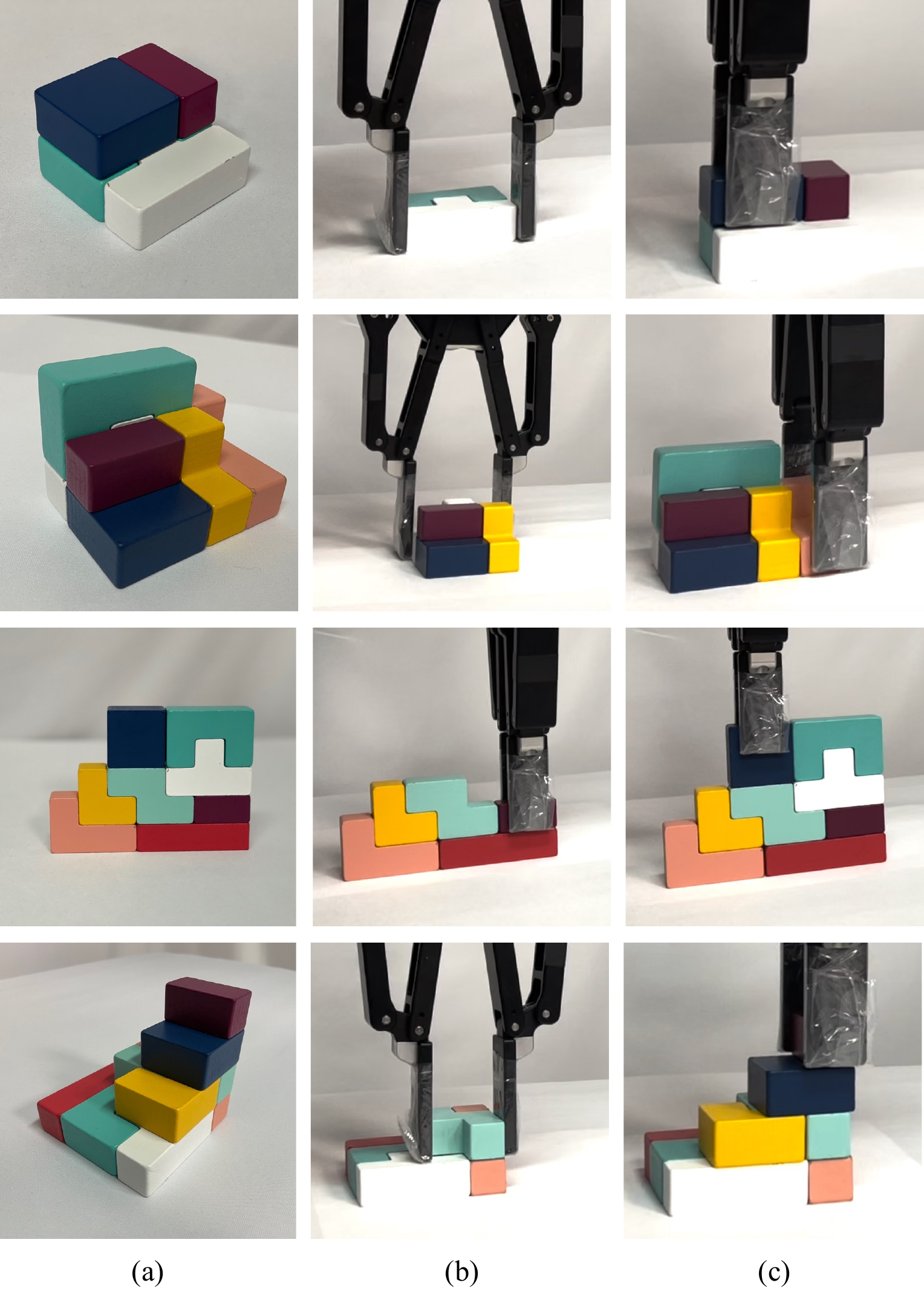}\vspace{-4mm}
	\caption{The results of robotic assembly. (a) Target structures. (b, c) The corresponding robotic assembly process.}
	\label{fig:assemble_result}
\end{figure}

\subsection{Robot Assembly Experiments}

The robotic system consists of a KUKA LBR iiwa R820 robot arm and a 2-finger Robotiq 2F-140 gripper, with an Intel RealSense D435i camera mounted to the gripper. 
The offset between the camera and gripper is obtained in an off-line calibration procedure. 
We conduct robot experiments leveraging Robotic Operating System (ROS). 
IIWA STACK~\cite{hennersperger2017towards}, a ROS metapackage for KUKA LBR iiwa robots, is employed to communicate between ROS and KUKA Sunrise software. 

We design four structures for assembly  (Fig.~\ref{fig:assemble_result}~(a)) and conduct 15 trials for each structure. 
As demonstrated in Table~\ref{tab:success_rate}, our methodology achieves a 100\% 2D detection rate and the 6D poses of all blocks are predicted, respectively. 
A trial is considered successful if all the blocks are properly assembled without severe collision and the assembly gap is less than 1mm as well. 
Among all the 60 trials, 52 are successful with the target structure constructed, achieving an 86.7\% success rate. 
To further analyze the whole construction process, we consider the perception, grasp, manipulation and assembly of a single block as a step. 
A trial consists of a different number of steps according to the number of blocks. 
We achieve a 93.5\% step success rate among all the steps in four structures. 

Noteworthy, the target structures are designed with tight tolerance and no gap is reserved between the blocks, thus only 1mm error is admissible during the assembly process. 
With a precise and robust 6D pose estimation algorithm, pose-guided 6D transformation strategy and 3-axis calibration, the assembly with tight tolerance is successfully implemented as shown in Fig.~\ref{fig:assemble_result}.  

\begin{figure}
	\centering
	\vspace{2mm}
	\includegraphics[width = 0.95\linewidth]{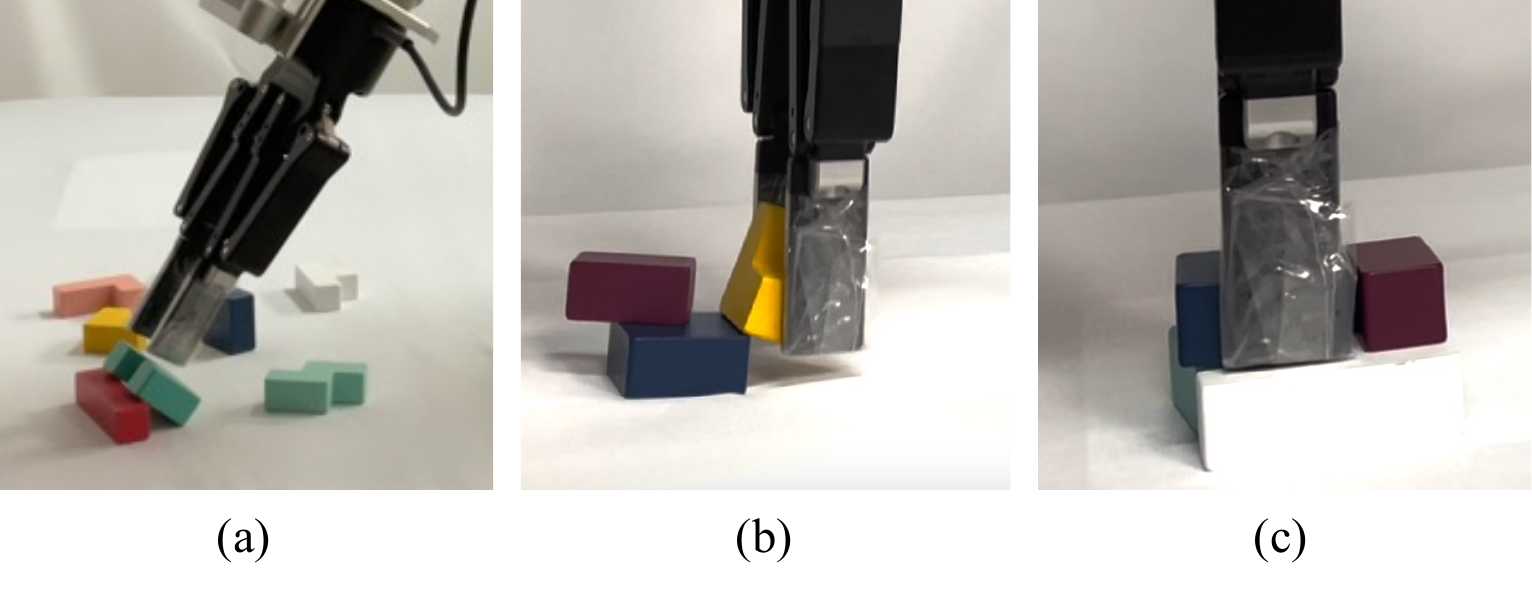}\vspace{-4mm}
	\caption{Failure cases. (a) caused by 6D pose estimation error. (b) caused by calibration error. (c) without 3-axis calibration. }
	\label{fig:failure_ablation}
\end{figure}

The system error is principally derived from 6D pose estimation and  calibration process. 
When the block is severely occluded or in certain ambitious conditions, our system may perceive a rough 6D pose. 
In most cases, the grasp can still be implemented and then the 6D pose error is eliminated by 3-axis calibration, which contributes to one successful assembly. 
Conversely, the 6D pose error that exceeds the tolerance of robotic grasping leads to a grasping failure (Fig.~\ref{fig:failure_ablation}~(a)). 

For the assembly error less than 1mm, only slight collisions and weeny gaps occur with no damage to the whole structure, while the assembly error larger than 1mm may wreck the whole structure (Fig.~\ref{fig:failure_ablation}~(b)). 
To summarize, our robotic assembly system is capable of assembling blocks with a tolerance of 1mm by RGB-only input. 

\begin{table}[t]
    \centering
    \caption{Results for robotic assembly task}
    \label{tab:success_rate}
    \setlength{\tabcolsep}{2.2mm}
    \begin{tabular}{ccccc}
    \hline
         Structure & \makecell{Number\\of Blocks} & \makecell{Detection\\Rate} & \makecell{Step\\Success Rate} & \makecell{Trial\\Success Rate}\\
    \hline
         1 & 4 & 100.0\% & 94.7\% & 93.3\% (14/15) \\
         2 & 6 & 100.0\% & 91.8\% & 86.7\% (13/15) \\
         3 & 8 & 100.0\% & 95.8\% & 86.7\% (13/15) \\
         4 & 8 & 100.0\% & 91.7\% & 80.0\% (12/15) \\
    \hline
         \multicolumn{2}{c}{Mean} & 100.0\% & 93.5\% & 86.7\% (52/60) \\
    \hline
    \end{tabular}
\end{table}

\subsection{Ablation Study on 3-axis Calibration}

To demonstrate the effectiveness of 3-axis calibration, we attempt to assemble structure 1, which consists of only four blocks, without exploiting the 3-axis calibration operation for the last 3 blocks.
Experimental results show that 13 out of 15 trials failed with severe collision and the collapse of the whole structure (Fig.~\ref{fig:failure_ablation}~(c)). 
Our 3-axis calibration operation outperforms direct assembly by a large margin, which indicates the indispensability of the disentanglement of 6D pose estimation and assembly process. 

It is worth noting that we make some assumptions in our approach, \textit{e.g.}, blocks with regular shapes, flat surfaces and stable flipping poses. 
Further exploration is needed for more general robotic assembly tasks. 
\section{CONCLUSION}

In this paper, we establish an integrated 6D robotic assembly system to assemble blocks with only RGB input. 
Leveraging a monocular 6D object pose estimation methodology, our system is capable of perceiving and interacting with the blocks with arbitrary initial poses. 
Abundant photorealistic training and validation data are generated exploiting physically-based rendering, which avoids complicated and labor-consuming 6D pose labeling in real scenes. 
Trained solely with synthetic data, our 2D detection network and 6D pose regression network are capable of directly transferring to real scenes without further training or using domain randomization techniques. 

Correspondingly, pose-guided 6D transformation along with collision-free assembly is proposed to deal with arbitrary initial poses and target poses. 
Our novel 3-axis calibration operation further promotes the precision and robustness of our system by decoupling 6D pose estimation and the assembly process. 
The experimental results demonstrate that our system is capable of robustly assembling blocks with 1mm tolerances.


\bibliographystyle{IEEEtran}
\bibliography{ref}

\end{document}